\documentclass{article}

\usepackage[final,nonatbib]{neurips_2023}

\usepackage[utf8]{inputenc} 
\usepackage[T1]{fontenc}    
\usepackage{hyperref}       
\usepackage{url}            
\usepackage{booktabs}       
\usepackage{amsfonts}       
\usepackage{nicefrac}       
\usepackage{microtype}      
\usepackage{xcolor}         
\usepackage{graphicx}
\usepackage{subcaption}
\usepackage{natbib}
\usepackage{appendix}

\usepackage{amsmath}
\usepackage{bbm}
\usepackage{enumitem}
\usepackage{pdflscape}

\newtheorem{condition}{Condition}

\DeclareMathOperator{\causal}{causal}

\title{
Monitoring the performance of machine learning algorithms that induce feedback loops: what is the causal estimand?
}

\author{%
  Jean Feng\thanks{jean.feng@ucsf.edu}$\hspace{0.15cm}^\ddagger$\\
\And Adarsh Subbaswamy\thanks{U.S. Food and Drug Administration, Center for Devices and Radiological Health}\\
\And Alexej Gossmann$^\dagger$
\And Harvineet Singh\thanks{University of California, San Francisco}
\And Berkman Sahiner$^\dagger$
\And Mi-Ok Kim$^\ddagger$
\And Gene Pennello$^\dagger$
\And Nicholas Petrick$^\dagger$
\And Romain Pirracchio$^\ddagger$
\And Fan Xia$^\ddagger$
}

\begin{document}

\maketitle

\vspace{-0.7cm}

\begin{abstract}
After a machine learning (ML)-based system is deployed, monitoring its performance is important to ensure the safety and effectiveness of the algorithm over time.
When an ML algorithm interacts with its environment, the algorithm can affect the data-generating mechanism and be a major source of bias when evaluating its standalone performance, an issue known as performativity.
Although prior work has shown how to \textit{validate} models in the presence of performativity using causal inference techniques, there has been little work on how to \textit{monitor} models in the presence of performativity.
Unlike the setting of model validation, there is much less agreement on which performance metrics to monitor.
Different monitoring criteria impact how interpretable the resulting test statistic is, what assumptions are needed for identifiability, and the speed of detection.
When this choice is further coupled with the decision to use observational versus interventional data, ML deployment teams are faced with a multitude of monitoring options.
The aim of this work is to highlight the relatively under-appreciated complexity of designing a monitoring strategy and how causal reasoning can provide a systematic framework for choosing between these options.
As a motivating example, we consider an ML-based risk prediction algorithm for predicting unplanned readmissions.
Bringing together tools from causal inference and statistical process control, we consider six monitoring procedures (three candidate monitoring criteria and two data sources) and investigate their operating characteristics in simulation studies.
Results from this case study emphasize the seemingly simple (and obvious) fact that \textit{not all monitoring systems are created equal}, which has real-world impacts on the design and documentation of ML monitoring systems.
\end{abstract}

\section{Introduction}

After a machine learning (ML)-based system is deployed, performance monitoring of the algorithm is necessary to minimize the release of misleading or outdated predictions \citep{Breck2017-qr, Finlayson2021-ad, Feng2022-mk}.
Nevertheless, compared to the topic of model development and pre-deployment evaluation, the topic of post-deployment performance monitoring remains relatively understudied \citep{Sculley2015-jw, Paleyes2022-od, Zhang2022-gu}.
Although prior works have suggested monitoring using sequential testing methods from statistical process control (SPC), these works largely assume an ``ideal data setting'' where the ML algorithm does not interact with its environment \citep{Montgomery2013-sk, Qiu2013-fa}.
However, ML algorithms are often designed to induce changes in their environment and can be a major source of bias.
This phenomenon, also known as performativity, algorithmic confounding, and feedback loops, has been discovered across various contexts including medicine, recommendation engines, and more \citep{Bottou2013-xp, Paxton2013-pv, Chaney2018-ak, Perdomo2020-cz}.

As a motivating example, consider a clinical risk prediction model for predicting 30-day unplanned readmission (see e.g. \citet{Horwitz2019-in}).
Many models have been developed to help hospitals reduce their readmission rates \citep{Burke2017-mz, Barbieri2020-ek, Mahmoudi2020-zn}, spurred by the Hospital Readmissions Reduction Program.
Prior works have documented various interventions hospitals may consider to reduce their readmission rates, such as post-discharge follow-ups.
Here we suppose that after the algorithm's predicted readmission risk for a patient is shown to the clinician, the clinician may choose between one of two treatments: scheduling a 5-day follow-up appointment or discharging with no plans for additional follow-up.
Performativity occurs when clinicians trust the algorithm, so that patients with high predicted risks tend to have high rates of follow-up.
Consequently, a procedure that directly estimates standalone performance using standard measures like AUC and accuracy without adjusting for (post-deployment) treatment propensities will be biased.
Prior work has shown how to obtain unbiased estimates of model performance, adjusted for performativity, using techniques from causal inference \citep{Sperrin2019-om}.
Nevertheless, there has been little discussion on how one should design an online monitoring procedure in the post-deployment setting.

One view of model monitoring is that it simply conducts model validation repeatedly over time, where the goal is to detect if the performance of a model has strayed from its performance at the time of initial model validation \citep{Nishida2007-lm, Klaise2020-fk, Schroder2022-rq, Corbin2023-gc}.
For instance, many models are evaluated in terms of their average performance, such as accuracy and AUC.
One may consider detecting changes in such quantities by combining causal inference with SPC and reweighting the monitoring data to match the target population \citep{Steiner2001-wj, Sun2014-yh, Cook2015-pd}.
However, there are many other ways to characterize model performance, including performance within particular subgroups \citep{Mitchell2021-wq} and conditional measures like a model's calibration curve \citep{Van_Calster2016-ey}.
Thus, monitoring mean performance is neither the only nor necessarily the best option.

A contrasting view is that the goal of model monitoring is not just to estimate changes in model performance but to minimize exposure to a faulty model.
This is particularly true in safety-critical settings like healthcare where elevated model error rates have severe consequences.
In such settings, a much stronger emphasis is placed on detection speed \citep{Montgomery2013-sk}.
Alerts produced by a monitoring procedure are handled by a quality assurance team, which investigates potential causes of performance decay and decides the most appropriate corrective action to take (e.g., updates to the model or data processing pipeline) \citep{Breck2017-qr, Raji2020-vx, Feng2022-mk}.

Monitoring procedures vary in their \textit{monitoring criterion} (i.e., what they aim to detect).
Taking a frequentist view in this paper, the monitoring criterion corresponds to a hypothesis test about a model's performance over time.
Monitoring criteria vary in how sensitive they are, their sample size requirements, and how interpretable their test statistics are.
In the presence of performativity (and other sources of bias), there are additional considerations, as the monitoring target must be cast within a causal framework to define the downstream effects of the ML algorithm: now our interest is in a \textit{causal} monitoring criterion, which impacts the types of data that can be used (observational versus interventional) and assumptions necessary for identifiability.
For instance, monitoring changes in AUC may be highly interpretable, but it is slow, as AUC is a relatively ``coarse'' metric \citep{Pencina2008-an}, and will require assumptions such as positivity in the observed data.
On the other hand, methods for monitoring algorithmic fairness and conditional performance measures can require weaker (or even no) assumptions regarding positivity, but may be less interpretable \citep{Feng2022-az}.

Although the importance of model monitoring has been recognized by model developers, users, and regulators alike \citep{Eaneff2020-ng, US_Food_and_Drug_Administration2021-tz, Schroder2022-rq}, previous discussions of model monitoring have lacked precision in terms of what the target estimand is, how it should be selected, and how it should be monitored.
The goal of this work is to demonstrate the value of causal reasoning when monitoring models in the presence of performativity, the underappreciated complexity of designing a monitoring strategy, and how different strategies can have vastly different consequences.
We do this by conducting an in-depth case study of the unplanned readmission model described earlier, though the conclusions in this work apply more broadly.
We walk through various procedures by considering different candidate monitoring criteria (Section~\ref{sec:criteria}) and candidate data sources (Section~\ref{sec:data}), and enumerate potential ways the readmission model can interact with its environment.
We then focus on the specific problem of ``interfering (medical) interventions'' (IMI) and compare the different monitoring strategies in an extensive simulation study (Section~\ref{sec:compare}).
Through this case study, we illustrate that \textit{not all monitoring systems are created equal}.
Designing an effective monitoring system requires a systematic causally-informed approach, and the (expected) operating characteristics of the deployed system should be transparent to users and other stakeholders.
Finally, we conclude with a discussion of open research problems at the intersection of causal inference and model monitoring.
Code is available at \url{https://github.com/jjfeng/monitoring_causally}.

\section{Related Work}
\label{sec:related}

Although the term ``performativity'' was recently coined in \citep{Perdomo2020-cz}, this phenomenon has been described across various problem domains.
One of the earliest types of performativity was perhaps described in \citep{Begg1983-bq, Zhou1998-ce, Alonzo2006-ke}: when assessing a diagnostic test, a well-known difficulty is that the test result itself can affect whether the patient receives a definitive assessment for disease status (e.g., a biopsy).
This problem, known as \textit{verification bias}, is typically viewed within a missing data framework.
More modern works focus on ML algorithms that aim to predict future outcomes but where their predictions can affect this very outcome \citep{Bottou2013-xp, Paxton2013-pv, Chaney2018-ak, Liley2021-yj}.
This type of performativity is typically formalized within a causal inference framework \citep{Mendler-Dunner2022-wk, Wu2022-wl}.
However, there are many similarities in the approaches taken, due to connections between missing data and causal inference \citep{Mohan2021-ip}.

Although there exist methods for one-time model evaluation in the presence of performativity \citep{Wu2022-wl}, the only work we are aware of that addresses monitoring in the presence of performativity is \citet{Feng2022-az}.
However, \citet{Feng2022-az} only describes one specific monitoring procedure.
\textit{The goal of this work is to provide a more global view of the monitoring landscape to describe the multitude of monitoring options one may consider in practice.}
More generally, there are few other works in the sequential process control literature that discuss the role of causal inference, as the field is primarily concerned with monitoring observational data quantities.
The most relevant methods to this work are those for monitoring differentially censored outcomes \citep{Steiner2001-wj, Steiner_2001-nw, Sun2014-yh}, again due to connections between missing data and causal inference.

\section{Background: sequential monitoring}
\begin{figure}
	\centering
	\vspace{-0.5cm}
	\includegraphics[width=0.9\textwidth]{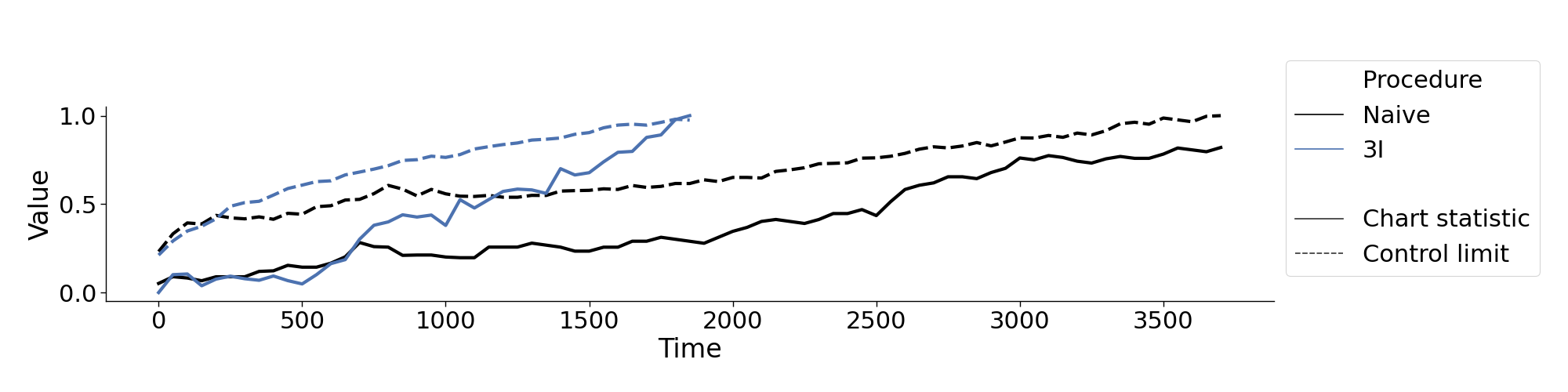}
	\vspace{-0.5cm}
	\caption{Example control charts, which plot the chart statistic (solid line) and control limit (dashed line) over time. When the chart statistics exceeds the control limit an alarm is fired.}
	\label{fig:control_charts}
\end{figure}

Given a monitoring criterion that defines how a data-generating process is expected to behave (also known as \textit{in-control} in the SPC literature), the goal of a sequential monitoring procedure is to detect when this process no longer behaves as expected (i.e. \textit{out-of-control}), given an incoming sequence of observations.
Formally, one can define the sequential monitoring problem as a hypothesis test, where the null hypothesis is that the process is in-control, and the sequential monitoring procedure as a sequential hypothesis testing procedure characterized by two functions:  chart statistic $\hat{C}: \{1,2,\cdots\} \to \mathbb{R}$ and control limit $h: \{1,2,\cdots\} \to \mathbb{R}$ \citep{Zhang2015-zk}.
When the chart statistic first exceeds the control limit, i.e., when $\hat{C}(t) > h(t)$ for some time $t$, an alarm is fired and the null hypothesis is rejected.
This is typically visualized via a control chart, which plots the two curves against each other (Figure~\ref{fig:control_charts}).

In the following sections, we explore different configurations for the three key ingredients of any monitoring system: (i) the monitoring criterion, (ii) the monitoring data, and (iii) the sequential hypothesis testing procedure.
We discuss the different options and their impacts on the operating characteristics of the overall monitoring procedure.
Here our interest is in procedures that maintain Type I error---the probability that it fires an alarm under the null for a given time span---and maximize power---the probability of firing an alarm after the process is out-of-control  \citep{Chu1996-fg, Zeileis2005-dc}.
One may also consider other operating characteristics depending on the problem setting \citep{Qiu2013-fa}.

\section{The case study: ML-based risk prediction algorithms}
\label{sec:example}

Our central case study is an algorithm for predicting patients' risk of unplanned readmission within 30 days.
Suppose the initial algorithm was evaluated based on its average positive predictive value (PPV) and negative predictive value (NPV).
The ML algorithm is allowed to evolve based on prior observations.
At each time point, we observe a new patient with variables $X_t$.
Upon querying the ML algorithm, the clinician is faced with two ``treatment options'': they may either schedule a 5-day follow-up appointment for that patient (intervention) or discharge them with no additional follow-up (standard of card, SOC).

\textbf{Notation.} Let $X_t$ represent variables for a patient drawn from the population at time $t$, uniformly at random.
$A_t$ is a binary ``treatment'' variable indicating whether the patient does or does not have a follow-up appointment ($a_t=1$ versus $a_t=0$, respectively).
Using potential outcomes notation, $Y_t(a_t)$ is a binary outcome which indicates if a 30-day unplanned readmission occurs under treatment option $a_t$, where $Y_t(a_t) = 1$ means the patient is readmitted.
For convenience, let $Y_t := Y_t(A_t)$ denote the observed outcome under the consistency assumption \citep{hernan2010causal}.
$\hat{f}_t(x, a)$ is the predicted risk from the ML algorithm at time $t$, given patient variables $x$ and treatment $a$.
In addition, let $\hat{y}_t$ be the binarized version of $\hat{f}_t$, e.g. $\hat{y}_t(x,a) = \mathbbm{1}\{\hat{f}_t(x,a) > b\}$ for some threshold $b$.
To formally describe how information accumulates over time, we define the process $\{(X_t, A_t, Y_t(0), Y_t(1), \hat{f}_t): t = 1,2,\cdots\}$ as being adapted to the filtration $\{\mathcal{F}_t: t = 1,2,\cdots\}$, where $\mathcal{F}_t$ is the sigma field generated by the collection of observations up to time $t$, i.e. $(\hat{f}_t, Y_{t-1}, A_{t-1}, X_{t-1}, Z_{t-1}, \cdots, Y_1, A_1, X_1, Z_1)$.

\subsection{Candidate monitoring criteria}
\label{sec:criteria}

When an ML system interacts with its environment to affect downstream outcomes, a causal framework is necessary to isolate the \textit{standalone} performance of the algorithm.
For instance, in this case study, the standalone PPV/NPV of the algorithm is
$\Pr\left(Y_t = \upsilon|\hat{y}_t(X_t,a) = \upsilon; \hat{f}_t \right)$ for $a,\upsilon \in \{0,1\}$.
Although one could try to approximate these measures using their observational versions, i.e.
$\Pr\left(Y_t = \upsilon|\hat{y}_t(X_t,A_t) = \upsilon; \hat{f}_t \right)$ for $a,\upsilon \in \{0,1\}$, the observational quantities are biased.
For instance, if clinicians tend to treat patients predicted to be at high risk of readmission, the observational PPV will be biased downwards while the observational NPV will be biased upwards.
As shown in later sections, this can lead to inflated Type I error and/or decreased power.
So rather than presenting typical monitoring criteria for observational quantities, we present \textit{causal} monitoring criteria.

In the following section, we describe three candidate criteria for monitoring model performance.
All candidate criteria are based on PPV/NPV, as the original algorithm was approved based on its overall PPV/NPV.
The criteria are listed from the most interpretable and least strict to the least interpretable and most strict, where stringency refers to strength in terms of model calibration \citep{Van_Calster2016-ey} and algorithmic fairness requirements.

\textbf{Criterion 1: The average PPV/NPVs should be maintained above some thresholds.}
For some set of predefined thresholds $\{c_{a \upsilon}: a,\upsilon \in\{0,1\}\}$, we consider the causal monitoring criterion
\begin{align}
	H_{0,C1}^{\causal} &:
	\Pr\left(Y_t(a) = \upsilon|\hat{y}_t(X_t,a) = \upsilon, \mathcal{F}_t \right) \ge c_{a \upsilon} \quad \forall t,a,\upsilon.
	\label{eq:hypo_npv}
\end{align}
The key benefit of this criterion is its interpretability, as it is a standard metric used in this case to approve the initial model.
This criterion is less strict in that the model's subgroup-specific performance can vary over time without any variation in its overall PPV/NPV.
(Conditioning on $\mathcal{F}_t$ in \eqref{eq:hypo_npv} lets us condition on the ML algorithm $\hat{f}_t$, which avoids the situation where $\hat{f}_t$ is a random variable that needs to be marginalized over.)

\textbf{Criterion 2: Subgroup-specific PPV/NPVs should be maintained above their respective thresholds.}
Motivated by concerns regarding algorithmic fairness, another criterion is that the PPV/NPV is maintained across predefined subgroups $\mathcal{S}_1,\cdots, \mathcal{S}_K$, such as those defined by protected attributes.
That is, for predefined subgroup-specific thresholds $\{c_{a\upsilon k}\}$, this criterion corresponds to checking the null hypothesis
\begin{align}
	H_{0,C2}^{\causal} &: \Pr\left(Y_t(a) = \upsilon|\hat{y}_t(X_t,a) = \upsilon, X_t\in \mathcal{S}_k , \mathcal{F}_t\right ) \ge c_{a\upsilon k} \quad \forall t,a,\upsilon, k.
	\label{eq:hypo_npv_subgroup}
\end{align}
This criterion is stricter than \eqref{eq:hypo_npv} and more sensitive to distribution shifts.
However, to decide the thresholds $c_{a\upsilon k}$, we may need to collect more data to estimate the PPV/NPV per subgroup, which may delay the time to model deployment.
Another concern is that any procedure that tests \eqref{eq:hypo_npv_subgroup} must account for multiple testing, so it is unclear how powerful a monitoring procedure for \eqref{eq:hypo_npv_subgroup} would ultimately be.

\textbf{Criterion 3: The predicted probabilities should not be over-confident for any subgroup.}
Finally, an even stricter criterion than checking subgroup-specific PPV/NPVs is to check that the algorithm's risk predictions are not overly extreme, in that the predicted risk should not be more extreme than the true adverse event rate for \textit{any} subgroup for some tolerance $\delta \ge 0$.
Formally, we check the null hypothesis:
\begin{align}
	H_{0,C3}^{\causal} &: \hat{s}_t(x, a)\left(\hat{f}_t(x, a) - \Pr(Y_t(a) = 1|x)\right) \le \delta \quad \forall t,a,x,
	\label{eq:hypo_score}
\end{align}
where $\hat{s}_t(x,a) = \text{sign}\left(\hat{f}_t(x, a) - 0.5 \right)$ is the predicted class assuming a threshold of 0.5.
This test can also be viewed as checking the strong calibration requirement, where the predicted probabilities are required to be within some distance of the true risk for each individual; models that satisfy the strong calibration requirement also satisfy strong notions of algorithmic fairness \citep{Van_Calster2016-ey, Hebert-Johnson2018-vb, Feng2023-ua}.
It is the strictest criterion among the three and, thus, the most sensitive to dataset shifts.
Although this can accelerate detection, the null hypothesis can be quite sensitive, particularly for small values of $\delta$, and may be violated only for a very small subgroup.
Thus, we must be careful when interpreting a rejection of \eqref{eq:hypo_score}, as it could indicate a decay in model performance or that the model was not well-calibrated to begin with.

\subsection{Data sources and causal models}
\label{sec:data}

To check monitoring criteria from Section~\ref{sec:criteria}, we must consider the types of data available and the biases they exhibit.
For this case study, suppose we can collect data from either an observational setting where healthcare providers are free to decide how to respond to the ML algorithm or an interventional setting where patients are randomized to the two treatment options.
As shown in Table~\ref{table:biases}, the number of potential biases can be large, and some may even interact.
Causal reasoning provides a framework for conceptualizing these biases and determining how best to adjust for them.
Stakeholders can also help identify which biases are less likely to occur or can be prevented through proper implementation of the ML system.

\begin{table}
	\centering
	\textbf{\textit{Potential sources of bias}}
	\vspace{-0.1cm}
	\begin{tabular}{p{5.8in}}
		\toprule
		\textbf{Interfering medical interventions}: Patients are scheduled for follow-up appointments with differing rates, driven in part by recommendations from the ML algorithm.
		\\
		\midrule
		\textbf{Selection bias}: The ML algorithm is only queried for a subset of patients \citep{Ladapo2013-ur, US_Food_and_Drug_Administration_undated-ir}, such as only the more difficult cases or subpopulations the algorithm is believed to perform well in. 
		\\
		\midrule
		\textbf{Off-label use}: ML algorithm may be queried in settings that are not recommended. For instance, the algorithm may be queried too early during an inpatient stay, or it might be queried during an outpatient visit even though it is only suggested for analyzing inpatient stays.
		\\
		\midrule
		\textbf{Patient trust}: The usage of the ML algorithm or its performance over time may motivate patients to leave or enter a hospital system \citep{Hashimoto2018-jl}.
		\\
		\midrule
		\textbf{Insurance coverage and billing}: The dollar amount covered by insurance companies for a given treatment may vary based on a patient's predicted risk, thereby modifying downstream medical decision making.
		\\
		\midrule
		\textbf{Circular definitions}: Hospitals may be more likely to readmit emergency department patients who were previously predicted to have high-risk of readmission.
	\end{tabular}
	\caption{
		Potential sources of bias when monitoring an ML algorithm for predicting risk of 30-day unplanned readmission.
		Many of these biases may vary over time, due to changes in how the clinician interacts with the algorithm and/or updates to the algorithm.
	}
	\label{table:biases}
\end{table}

\begin{figure}
	\centering
	\vspace{-1cm}
	\includegraphics[width=0.6\textwidth]{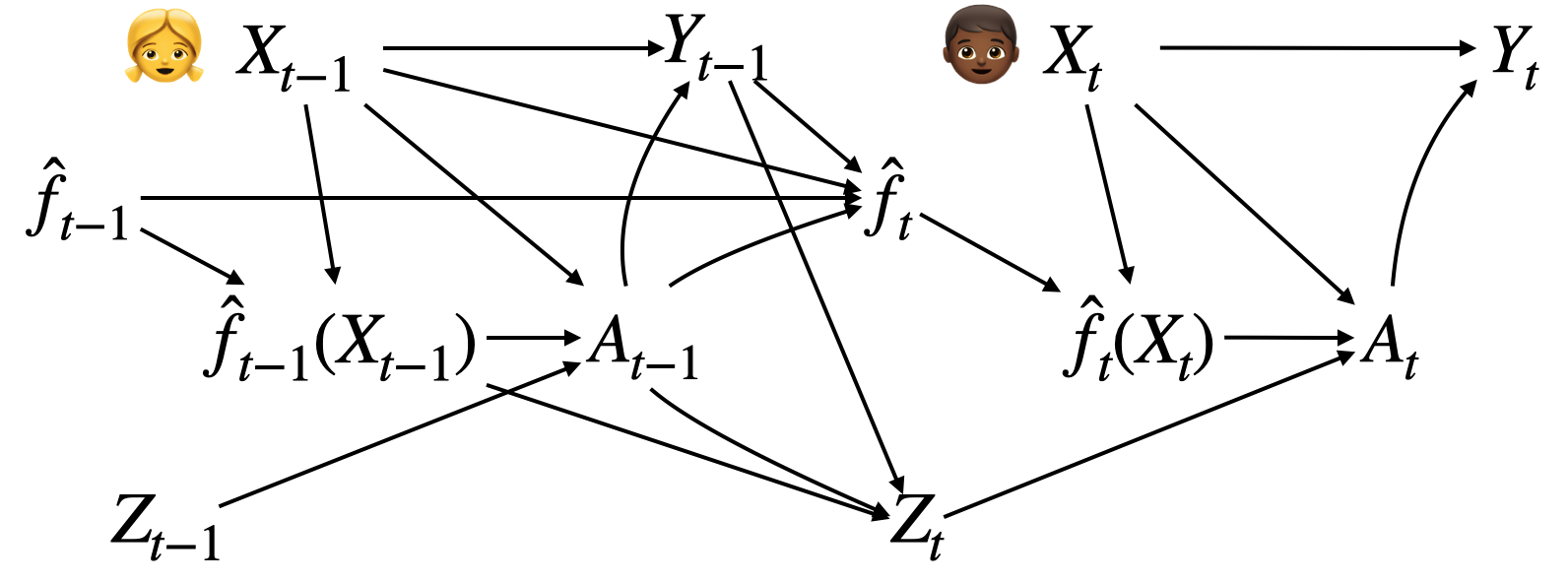}
	\vspace{-0.4cm}
	\caption{Causal model describing interfering medical interventions induced by an evolving ML-based risk prediction algorithm $\hat{f}_t$.
		$X_t$ denotes variables for the patient being queried for at time $t$.
		$Z_t$ denotes non-patient variables that may affect treatment decisions, such as past performance of the ML algorithm.
		$A_t$ denotes treatment assignment and $Y_t$ denotes the patient's outcome.
		Note that $\hat{f}_t(X_t) = (\hat{f}_t(X_t,0), \hat{f}_t(X_t,1))$.
	}
	\label{fig:dag}
\end{figure}

For simplicity, suppose we are mainly concerned with the problem of ``interfering (medical) interventions'' (IMI) \citep{Paxton2013-pv, Dyagilev2016-ii, Lenert2019-uf, Liley2021-yj}.\footnote{This problem has been previously referred to as ``confounding medical interventions.'' However, strictly speaking, predictions from the ML algorithm do not confound the relationship between $A$ and $Y$. Hence, we have modified the name of this phenomenon in this paper.}
IMI is the problem where the ML algorithm modifies treatment patterns, resulting in dependent ``censoring'' of counterfactual outcomes.
For instance, if we estimated the NPV of $\hat{y}_{t}(\cdot, a)$ by directly calculating the NPV among patients who received treatment $a$, this estimate would be biased upwards because patients who are assigned treatment $a$ will tend to have lower risk if the algorithm is performing as expected.
As illustrated in the simulation study in Section~\ref{sec:compare}, this can lead to unnecessarily long detection delays.

To define the data-generating mechanism for the observational setting more formally, we assume data follows the directed acyclic graph (DAG) shown in Figure~\ref{fig:dag}.
IMI is caused by the arrows entering the treatment decision $A_t$.
More specifically, we assume that $A_t$ depends on patient variables $X_t$, output from the ML algorithm $\hat{f}_t(X_t, \cdot)$, and some other random variable $Z_t$, which represents non-patient factors that may influence a clinician's decision to treat.
For instance, $Z_t$ may be the performance of the ML algorithm recently.

\subsection{Candidate monitoring strategies}
\label{sec:options}

We now describe various sequential procedures for detecting violations of the monitoring criteria listed in Section~\ref{sec:criteria}.
This section presents variants of the popular CUSUM control chart \citep{Page1954-gr} for each candidate criteria and data source.
The methods vary in what identifiability assumptions they require and how the chart statistics are defined.
The Appendix describes a unified procedure for constructing control limits that control the Type I error and provides detailed identification assumptions.

\subsubsection{Monitoring the average PPV/NPV (Criterion 1)}

To motivate the chart statistic for monitoring \eqref{eq:hypo_npv}, let us first suppose the counterfactual outcomes are observed.
Over the time window from $t_1$ to $t_2$, the difference between the threshold and the true PPV/NPV can be estimated by
\begin{align}
	c_{a \upsilon}- \frac{\sum_{i=t_1}^{t_2} \mathbbm{1}\{Y_i(a) = \upsilon \} \mathbbm{1}\{\hat{y}_i(X_i,a) = \upsilon\}}
	{\sum_{i=t_1}^{t_2} \mathbbm{1}\{\hat{y}_i(X_i,a) = \upsilon \}}
	\quad \text{for } a,\upsilon = 0,1.
\end{align}
Equivalently, we can rescale by the denominator and monitor $\Pr(Y_i(a) = \upsilon, \hat{y}_i(X_i,a) = \upsilon|\mathcal{F}_i)$ per
\begin{align}
	\frac{1}{t_2 - t_1+ 1}\sum_{i=t_1}^{t_2} \left(c_{a\upsilon} - \mathbbm{1}\{Y_i(a) = \upsilon \}\right) \mathbbm{1}\{\hat{y}_i(X_i,a) = \upsilon\}.
\end{align}
As the actual change time is unknown, the CUSUM procedure \citep{Page1954-gr} defines a scan statistic that calculates the maximum \underline{cu}mulative \underline{sum} of residuals over all possible changepoints:
\begin{align}
	\hat{C}^{\causal}(t) &= \max_{a,\tau,\upsilon} \sum_{i=\tau}^{t}
	(c_{a \upsilon} - \mathbbm{1}\{Y_i(a) = \upsilon \}) \mathbbm{1}\{\hat{y}_i(X_i,a) = \upsilon \}.
	\label{eq:cusum_traditional}
\end{align}
Under the null, the conditional mean of each summand in \eqref{eq:cusum_traditional} is zero, which implies that the CUSUM chart statistic converges to a Brownian motion; this facilitates the construction of control limits that control the Type I error  \citep{Chu1996-fg, Zeileis2005-dc}.
Under the alternative, the conditional mean of each summand is positive, so \eqref{eq:cusum_traditional} grows at an $O_p(t)$ rate and has an asymptotic power of one.

\paragraph{Option 1N: A na\"ive monitoring procedure}
If one was to ignore the impact of IMI, one may choose to directly monitor the average PPV/NPV in the observed data using the chart statistic
\begin{align}
	\hat{C}^{1\texttt{N}}(t) &= \max_{a,\tau, \upsilon} \sum_{i=\tau}^{t}
	(c_{a \upsilon} - \mathbbm{1}\{Y_i = \upsilon\}) \mathbbm{1}\{\hat{y}_i(X_i,A_i) = \upsilon, A_i=a\}.
	\label{eq:naive_cusum}
\end{align}
However, it is clear from the causal model that this quantity can be biased upwards or downwards depending on the treatment propensities. Thus, the alarm rate for such a procedure may be too high or too low.

\paragraph{Option 1I (Interventional)}
To avoid biases introduced by IMI, one option is to randomize treatment with known weights $p_i(a|x,z,\hat{f}) \in (0,1)$.
By design, we have through inverse propensity weighting (IPW)
\begin{align}
	\hspace{-0.7cm}
	\mathbbm{E}\left[\left(c_{a \upsilon} - \frac{\mathbbm{1}\{Y_i =\upsilon, A_i = a\}}{{p}_i(A_i|X_i,Z_i, \hat{f}_i)}\right) \mathbbm{1}\left\{\hat{y}_i(X_i,a) = \upsilon \right\}\right]
	= 
	\mathbbm{E}\left[\left(c_{a \upsilon} - \mathbbm{1}\{Y_i(a) =\upsilon\}\right) \mathbbm{1}\left\{\hat{y}_i(X_i,a) = \upsilon\right\} \right],
	\label{eq:cond_exch}
\end{align}
without needing to make any additional identifiability assumptions.
Consequently, we can monitor the CUSUM chart statistic with IPW
\begin{align}
	\hat{C}^{1\texttt{I}}(t) &= \max_{\tau,a,\upsilon} \sum_{i=\tau}^{t}
	\left(c_{a \upsilon} - \frac{\mathbbm{1}\{Y_i = \upsilon, A_i = a\}}{p_i(A_i|X_i,Z_i, \hat{f}_i)}\right) \mathbbm{1}\{\hat{y}_i(X_i,a) = \upsilon\}.
	\label{eq:randomize}
\end{align}

\paragraph{Option 1O (Observational)}
Let us continue with the above example, but we do not randomize treatment this time.
Although \eqref{eq:cond_exch} no longer holds in general, it does hold if we assume positivity---i.e. $p_t(A_t=1 | X_t, Z_t, \hat{f}_t) \in (0,1)$ almost everywhere for the oracle model $p_t$---and (sequential) conditional exchangeability---i.e. $Y_t(a) \perp A_t | X_t, Z_t, \mathcal{F}_t$---for all times $t$. 
Under these identifiability assumptions, we may extend \eqref{eq:randomize} to the observational setting via a two-part solution similar to that in \citet{Steiner2001-wj}, where we (i) monitor PPV/NPVs assuming the propensity model is constant and (ii) monitor for changes in the propensity model.
As the treatment propensities are unknown, we approximate \eqref{eq:randomize} using estimates of the propensities instead.
That is, letting $\hat{p}_t$ denote our estimate of the propensity model at time $t$, the chart statistic is defined as
\begin{align}
	\hat{C}^{1\texttt{O}}_a(t) &= \max_{\tau,a,\upsilon} \sum_{i=\tau}^{t}
	\left(c_{a \upsilon} - \frac{\mathbbm{1}\{Y_i = \upsilon, A_i = a\}}{\hat{p}_i(A_i|X_i,Z_i, \hat{f}_i)}\right) \mathbbm{1}\{\hat{y}_i(X_i,a) = \upsilon\}.
	\label{eq:ipw}
\end{align}
Because \eqref{eq:ipw} requires an estimate of the propensity model at $t = 0$, this monitoring procedure requires conducting a ``pre-monitoring study'' after the deployment of the algorithm, where the sole purpose is to learn treatment assignment patterns \citep{Feng2022-az}.
During this phase, healthcare providers are asked to make treatment decisions based on predictions from the ML algorithm, but we only begin monitoring upon completion of this phase.
(This pre-monitoring study is akin to Phase I in SPC \citep{Qiu2013-fa}.)
Another drawback of this procedure is that we cannot verify the identifiability assumptions \citep{Pearl2016-zf}.
In fact, violations of the positivity condition are more likely to occur the more accurate the ML algorithm is \citep{Lenert2019-uf}.

\subsubsection{Monitoring subgroup-specific PPV/NPVs (Criterion 2)}

We now extend the procedures from above to monitor subgroup-specific PPV/NPVs instead.
For the interventional setting (\textbf{Option 2I}), we define the chart statistic as the maximum of \eqref{eq:randomize} calculated for each subgroup, i.e.
\begin{align}
	\hspace{-0.2in}
	\hat{C}^{2\texttt{I}}(t) &= \max_{\tau,a,\upsilon,k} w_{k \upsilon}
	\sum_{i=\tau}^t
	\left(
	c_{a\upsilon k}
	- \frac{\mathbbm{1}\{Y_i = \upsilon, A_i = a\}}{p_i(A_i|X_i,Z_i, \hat{f}_i)}\right) \mathbbm{1}\{\hat{y}_i(X_i,a) = \upsilon, X_i \in \mathcal{S}_k\}.
	\label{eq:subgroup_chart}
\end{align}
where $w_{k \upsilon}$ is the weight associated with subgroup $\mathcal{S}_k$ and label $\upsilon$.
For the observational setting (\textbf{Option 2O}), the chart statistic $\hat{C}^{2\texttt{O}}$ is exactly the same as $\hat{C}^{2\texttt{I}}$ except that the oracle propensity model is replaced by its estimate.

Compared to tests for Criterion 1, these tests can be more powerful when performance decay is confined to only one of the specified subgroups.
Also, in the observational setting, the major benefit of Option 2O over Option 1O is that the subgroup weights $w_{k \upsilon}$ can be selected to downweight subgroups with (near-)violations of the positivity condition; we can even remove such subgroups altogether.
For instance, the empirical analyses in Section~\ref{sec:compare} set $w_{k \upsilon}$ to be the inverse of the estimated standard deviation of the summand in \eqref{eq:subgroup_chart} for subgroup $\mathcal{S}_k$, with respect to the pre-change distribution. This way, the variance of the CUSUM statistic in each subgroup has roughly the same variance.

Potential concerns of these procedures are that: (i) their power depends on the overlap between $\{\mathcal{S}_k\}$ and the actual subgroup that experiences performance decay, (ii) more data needs to be collected upfront to estimate the initial PPV/NPV per subgroup and select the subgroup-specific thresholds $c_{ak}$, and (iii) we must perform multiplicity correction both across subgroups and over time to control the Type I error.

\subsubsection{Checking for over-confident risk predictions (Criterion 3)}
To test \eqref{eq:hypo_score}, we note that the expected residual
\begin{align}
	\mathbbm{E}\left[
	\hat{s}_t(X_t, a) \left(\hat{f}_t(X_t,a) - Y_t(a)\right) - \delta \mid X_t=x, \mathcal{F}_t 
	\right]
	\label{eq:causal_resid}
\end{align}
is no larger than zero for almost every $x$ under the null.
On the other hand, \eqref{eq:causal_resid} will take on positive values at some values of $x$ under the alternative.
So to check for violations of \eqref{eq:hypo_score}, one approach is to check if \eqref{eq:causal_resid} when averaged over \textit{any} marginal distribution of ${X}_t \mid \mathcal{F}_t$ is large.
In particular, \textbf{Option 3I} monitors the following chart statistic in the interventional setting:
\begin{align}
	\hat{C}^{3\texttt{I}}(t) &= \max_{\tau,k} w_{k \upsilon}
	\sum_{i=\tau}^t
	\left[\hat{s}_t(X_t, a) \left (\hat{f}_i(X_i, A_i) - Y_i \right) - \delta\right] \mathbbm{1}\{X_i \in \mathcal{S}_k\}.
	\label{eq:chart_score}
\end{align}
Compared to Options 1I and 2I, a major benefit of this approach is that there are no inverse weights, so it does not require the positivity assumption.
However, the drawback is that \eqref{eq:chart_score} does not monitor a standard performance metric and is thus not very interpretable.
Nevertheless, it may still be effective as a monitoring metric.

Another drawback of this approach is its sensitivity to miscalibration in the model, even in small subgroups.
In addition, this procedure places more weight on checking predictions for treatments with higher propensities.
So compared to the procedures with inverse weights, it will have lower power when distribution shifts concentrate in low-propensity regions.

In the observational setting, \textbf{Option 3O} assumes sequential conditional exchangeability so that \eqref{eq:causal_resid} is equal to the observational quantity $\mathbbm{E}[\hat{s}_t(X_t,A_t)(Y_t - \hat{f}_t(X_t,A_t)) - \delta \mid A_t=a, X_t=x, \mathcal{F}_t]$ for almost every $(x,a)$.
As such, the chart statistic $\hat{C}^{3\texttt{O}}$ is defined exactly the same as $\hat{C}^{3\texttt{I}}$.
Compared to Options 1O and 2O, the benefits of this approach are that: (i) there is no need to model the propensity and the propensity model may even vary over time, (ii) we do not need to conduct a pre-monitoring study (as opposed to Option 2O), and (iii) the positivity assumption is not needed to control the Type I error.

\subsection{Comparing candidate strategies: a simulation study}
\label{sec:compare}
To design the most suitable monitoring solution, one needs to compare the multitude of monitoring options with respect to various dimensions, including their interpretability, fairness, data requirements, identification assumptions, hyperparameters, and operating characteristics.
Many of these properties can be summarized using a table, as shown in Table~\ref{tab:pro_con}.
Given that we do not know how exactly the data will evolve over time, the monitoring procedures should be evaluated across a variety of simulated data settings; active stakeholder engagement is necessary to ensure this set of simulations is sufficiently comprehensive.
In this section, we illustrate how such simulations can illuminate differences in the operating characteristics between the aforementioned monitoring options.

\begin{table}
	\hspace{-0.5in}
	\begin{tabular}{c|c|c|p{1.7in}|p{1.5in}|p{1.2in}}
		Procedure & Interpretability & Fairness &  Data requirements & Assumptions & Hyperparameters \\
		\toprule
		1I & High & None & Interventional & Positivity & None \\
		\midrule
		1O & High & None &Observational, Must conduct pre-monitoring phase & Positivity, Conditional Exchangeability  & None \\
		\midrule
		2I & High & Moderate &Interventional & Positivity & Subgroups, subgroup PPV/NPV \\
		\midrule
		2O & High & Moderate &Observational, Must conduct pre-monitoring phase & Positivity, Conditional Exchangeability & Subgroups, subgroup PPV/NPV \\
		\midrule
		3I & Medium & Strong & Interventional & None& Subgroups, tolerance level \\
		\midrule
		3O & Medium & Strong & Observational, No pre-monitoring phase & Conditional Exchangeability & Subgroups, tolerance level
	\end{tabular}
	\caption{Properties of different monitoring procedures}
	\label{tab:pro_con}
\end{table}

For this case study, suppose we believe healthcare providers will closely follow the ML algorithm's recommendations in the observational setting.
We compare against an interventional setting where randomization weights are defined using a logistic regression model that favors the recommended treatment but with less extreme propensities.
An alternative view of this comparison is that it illustrates how different levels of clinician trust can impact detection delay.
We simulate a sudden shift in the conditional distribution at time $t=500$ and vary the following factors:
\begin{itemize}
	\item Whether the conditional distribution of $Y(X,a)$ shifts for treatment $a=0$ versus $a=1$. Treatment $a=0$ is assigned at a higher rate in the population in the observational setting. (\texttt{Treatment A=0} vs \texttt{A=1 Shift})
	\item The subpopulation experiencing calibration decay. We simulate a shift in the conditional distribution of $Y(a)$ for all $X$ (\texttt{Subgroup all}), a known subgroup with prevalence of 40\% (\texttt{Subgroup known}), and a misspecified subgroup with prevalence of 35\% (\texttt{Subgroup misspec}).
	\item How much the risk $\Pr(Y(X,a)=1|X)$ increased (\texttt{Magnitude 10\%} versus \texttt{20\%}).
\end{itemize}
In the Appendix, we also present results when the simulated distribution shift is gradual.

For all the monitoring criteria, the null hypotheses allow for only a 2\% drop in the PPV/NPV or 2\% difference in the true risk.
As such, none of the null hypotheses are true.
The control limit for each procedure controls the Type I error rate over 4000 time points at 10\%.
We implement procedures for Criteria 2 and 3 to monitor shifts with respect to all patients, the known subgroup, and the complement of the known subgroup.
Full simulation details are provided in the Appendix.

As shown in Figure~\ref{fig:cusum}, there are substantial differences in power across the twelve simulation settings.
The performance of the na\"ive procedure was highly erratic because it fails to adjust for IMI.
It attains the fastest time to detection when there is a shift in the entire population, because the PPV drops substantially once the ML algorithm is deployed and even more after the shift.
However, the na\"ive procedure can be one of the slowest detectors in other scenarios.
Among the methods that correctly adjust for IMI, monitoring criterion 3 was consistently the most powerful approach.
The gap in performance between criterion 3 versus 1 and 2 tended to be larger when the magnitude and/or prevalence of performance decay was smaller.
The ranking between methods for monitoring criterion 1 and 2 alternated based on how widespread calibration decay was and whether it was contained within a known subgroup.

Interestingly, collecting interventional data was beneficial only in certain circumstances.
In particular, it led to faster detection when distribution shifted with respect to treatment $A=1$ but slower detection when the shift was with respect to treatment $A=0$.
This is because the rate of assigning treatment $A=0$ was higher in the observational setting.
Randomization tended to be more helpful for criteria 1 and 2 than 3, because the inverse propensity weights became less extreme.
In contrast, detection speed and power using Options 3O versus 3I were generally quite similar.

By conducting a thorough analysis of various monitoring options, we can weigh the utility of various procedures.
In practice, one would likely simulate numerous other data settings in addition to those shown. For instance, one may investigate the impact of changes in clinician trust over time, model updating over time, violations of the assumptions made by the monitoring procedures, and more.
For this case study, Table~\ref{tab:pro_con} and the simulation results suggest that Option 3O is a reasonable monitoring strategy.
This would significantly simplify the deployment of the ML algorithm, though it may require recalibration of the initial model to ensure that it is strongly calibrated.

\begin{figure}
	\centering
	
	\textit{Treatment A=0 shift, Magnitude 10\%}
	
	\footnotesize{\textit{\qquad \qquad Shifts in All \qquad\qquad \qquad \qquad  Subgroup known  \qquad  \qquad \qquad\qquad  Subgroup misspec}}
	\includegraphics[width=0.3\textwidth]{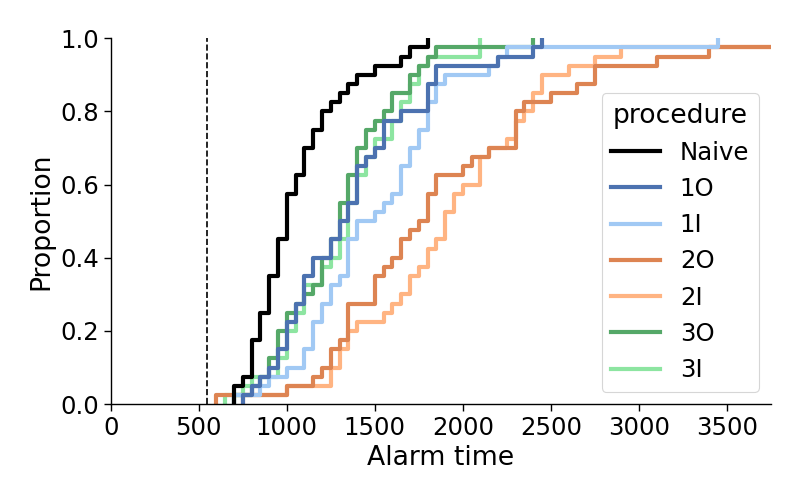}
	\includegraphics[width=0.3\textwidth]{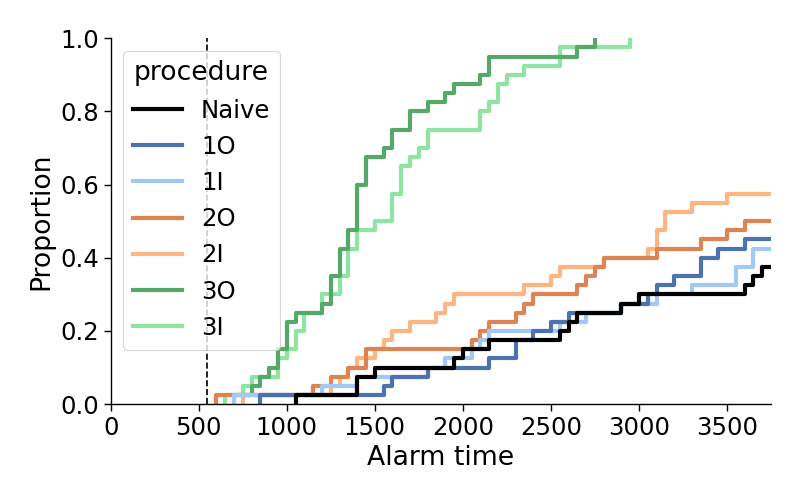}
	\includegraphics[width=0.3\textwidth]{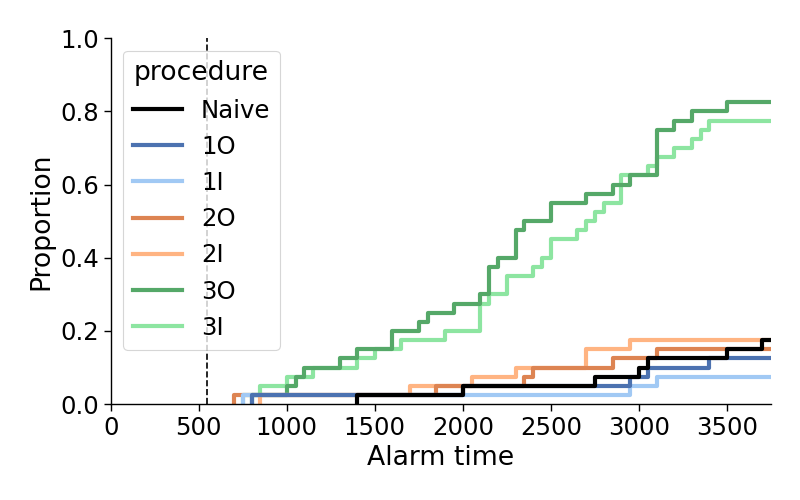}
	
	\textit{Treatment A=1 shift, Magnitude 10\%}
	
	\footnotesize{\textit{\qquad \qquad Shifts in All \qquad\qquad \qquad \qquad  Subgroup known  \qquad  \qquad \qquad\qquad  Subgroup misspec}}    \includegraphics[width=0.3\textwidth]{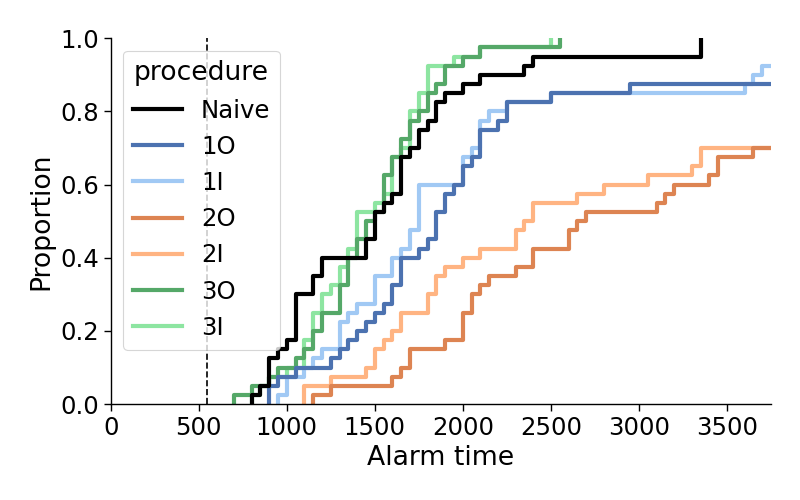}
	\includegraphics[width=0.3\textwidth]{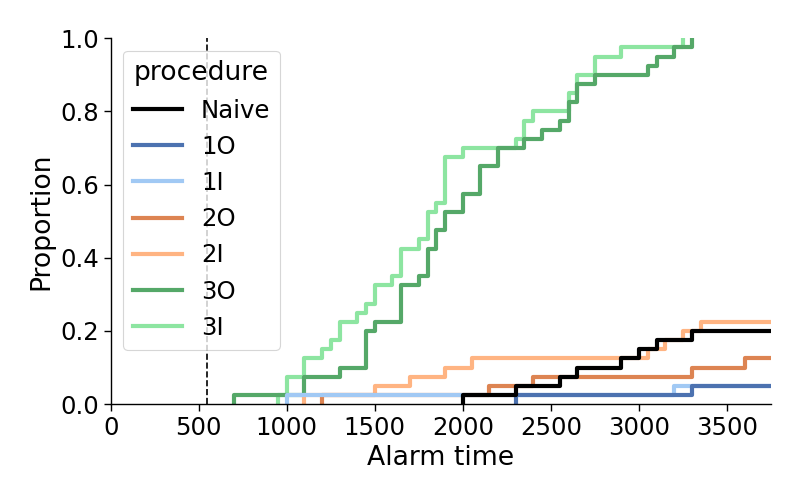}
	\includegraphics[width=0.3\textwidth]{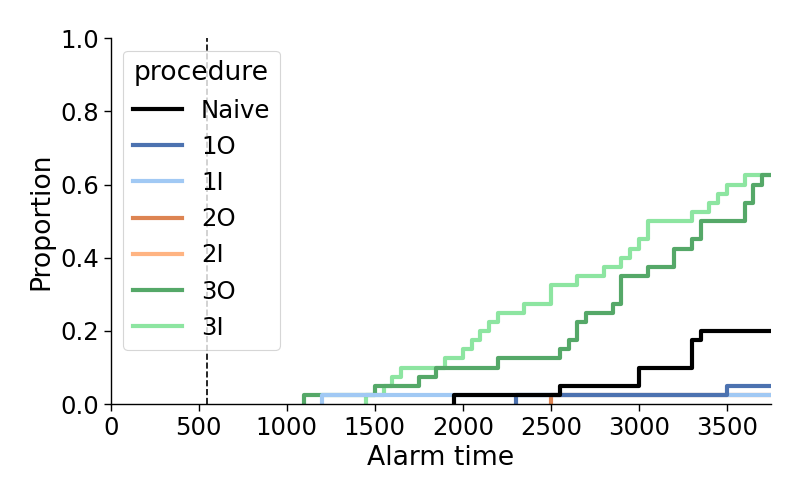}
	
	\textit{Treatment A=0 shift, Magnitude 20\%}
	
	\footnotesize{\textit{\qquad \qquad Shifts in All \qquad\qquad \qquad \qquad  Subgroup known  \qquad  \qquad \qquad\qquad  Subgroup misspec}}
	\includegraphics[width=0.3\textwidth]{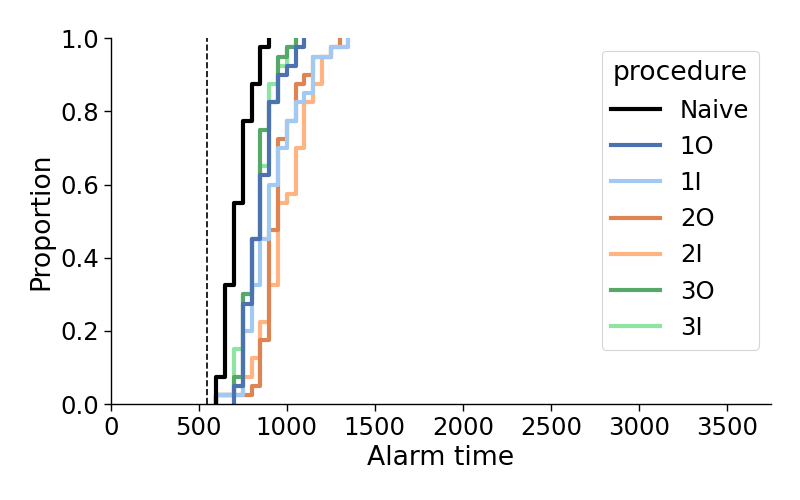}
	\includegraphics[width=0.3\textwidth]{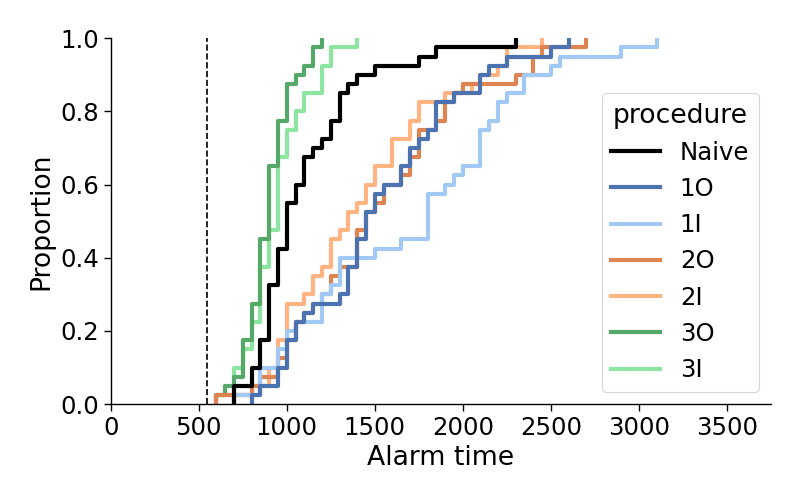}
	\includegraphics[width=0.3\textwidth]{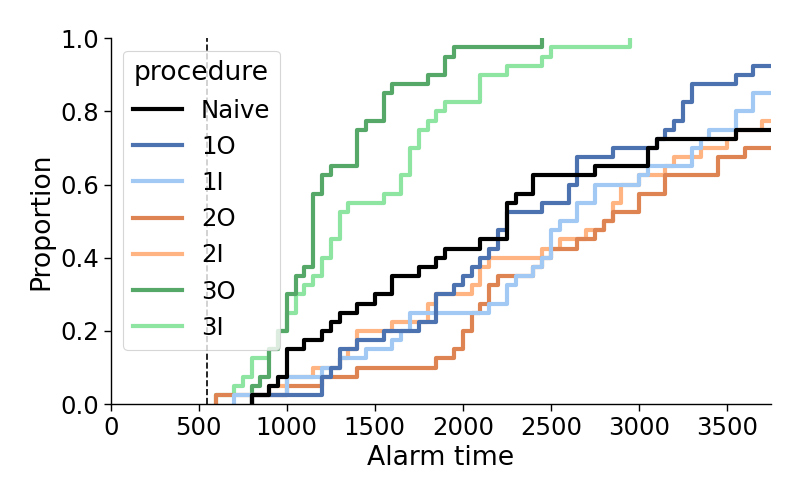}
	
	\textit{Treatment A=1 shift, Magnitude 20\%}
	
	\footnotesize{\textit{\qquad \qquad Shifts in All \qquad\qquad \qquad \qquad  Subgroup known  \qquad  \qquad \qquad\qquad  Subgroup misspec}}
	\includegraphics[width=0.3\textwidth]{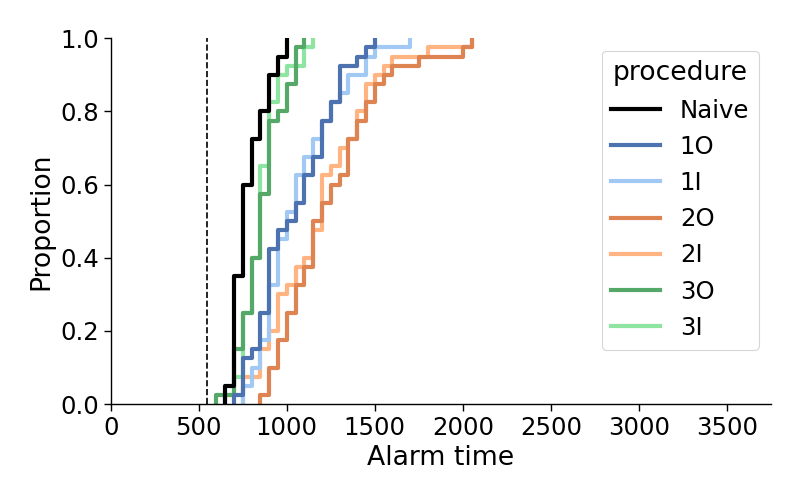}
	\includegraphics[width=0.3\textwidth]{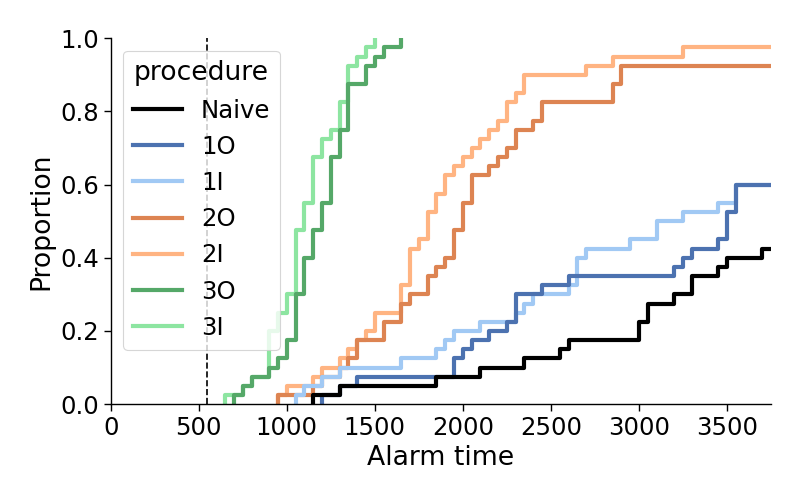}
	\includegraphics[width=0.3\textwidth]{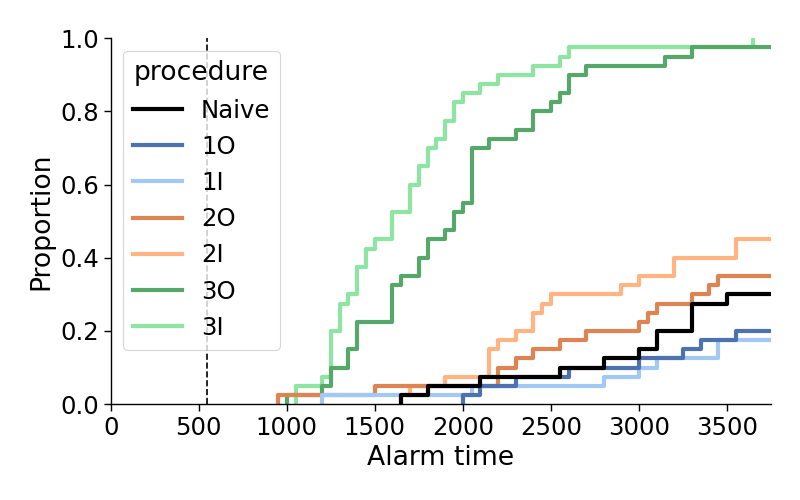}

	\caption{Statistical power of different procedures, as characterized by the proportion of alarms fired at each time point. Dashed vertical line is the time of the distribution shift.}
	\label{fig:cusum}
\end{figure}

\section{Discussion}

Although the problem of performativity complicates monitoring the performance of ML algorithms, we show in this work that causal inference provides a way to conceptualize biases induced by ML algorithms.
However, the question when designing a monitoring strategy is not only how to adjust for performativity but also \textit{what} we should be monitoring.
Not all monitoring systems are created equal.
There are a multitude of monitoring strategies to choose from, that vary in their data source, identifiability assumptions, interpretability of their test statistics, the number of hyperparameters, and more.
For instance, we find in this case study that checking for fairness violations can be a powerful approach to detecting model decay early, revealing an interesting connection between algorithmic fairness and performance monitoring.
More generally, ML quality teams should conduct systematic evaluations and seek input from diverse stakeholders to choose between various options, as the choice is often not clear upfront.
After a monitoring system has been put in place, documentation should also be available to users to understand the operating characteristics of the monitoring procedures.

There are still many open areas for research, many of which we believe can be answered with the help of causal inference.
First, the interplay between model monitoring and other types of performativity warrants further investigation, including those highlighted in Table~\ref{table:biases} and Section~\ref{sec:related}.
Moreover, with the proliferation of ML systems, it becomes increasingly important to study interactions between multiple ML algorithms.
Second, different types of experimental designs should be considered.
In this work, we only consider collecting data from fully observational or fully interventional settings.
One may also consider combining both data sources \citep{Bareinboim2016-im}, adaptive randomization \citep{Pallmann2018-wy}, or even quasi-experimental designs such as encouragement designs \citep{Hirano2000-ce} and instrumental variable analyses \citep{Baiocchi2014-bz}.
More generally, one can even consider other types of estimands.
We have focused on the standalone performance of the ML algorithm, but one may also be interested in the overall effect of the device on patient outcomes, which would require different randomization schemes \citep{Bossuyt2000-ft, FDA2022-ny}.
Finally, this work studies the performance of variants of the CUSUM, but future work should consider other types of monitoring procedures including those based on anytime inference \citep{Grunwald2019-wr, Shekhar2023-pz} and Bayesian inference \citep{West1997-fa}.



\section*{Acknowledgments}

This work was funded through a Patient-Centered Outcomes Research Institute® (PCORI®) Award (ME-2022C1-25619).
The views presented in this work are solely the responsibility of the author(s) and do not necessarily represent the views of the PCORI®, its Board of Governors or Methodology Committee, and the Food and Drug Administration.

\bibliographystyle{plainnat}
\bibliography{main}

\appendix




\section{Calculating control limits}

For the procedures monitoring Criterion 3, we use the Monte Carlo procedure described in \citep{Zhang2015-zk, Feng2023-ua} to construct dynamic control limits (DCL).
At each time $t$, we bootstrap $Y_t^*$ given $X_t$ under the worst-case null hypothesis satisfying \eqref{eq:hypo_score}, which corresponds to the case where $\Pr(Y_t^* = 1|X_t, \hat{f}_t) = \hat{f}_t(X_t, A_t) + \delta$.
For each of the bootstrap sequences, we calculate the corresponding chart statistics.
The DCL is then chosen to satisfy an alpha spending function.
Here, we use an alpha spending function that uniformly spends the $\alpha$ over time.
This procedure provably controls the Type I error in finite samples \citep{Feng2023-ua}.

In general, Criteria 1 and 2 are weaker than 3.
To conduct a more fair comparison in the simulation studies, we construct DCLs for their corresponding test statistics to control the Type I error rate under Criteria 3.
Thus the sensitivity of the methods for monitoring Criteria 1 or 2 are actually higher than they would be otherwise.
Also, for illustrative purposes, we make the simplifying assumption in the simulation studies that the estimation error for the propensity model is negligible; this is true, for instance, if we had a sufficiently long pre-monitoring phase.
To formally adjust for estimation error, one can use ideas such as that in \citep{Gombay2002-fe, Dette2020-mr, Feng2022-az}.

\section{Identification assumptions}
The identification assumptions needed for procedures 1I and 2I are as follows:
\begin{condition}[Positivity]
	For some $\epsilon > 0$, weights $p_t(a_t|x_t,z_t,\hat{f}_t)$ satisfy
	$p_t(a_t|x_t,z_t,\hat{f}_t) \in (\epsilon,1-\epsilon)$
	for all time points $t$.
\end{condition}
\begin{condition}[Conditional Exchangeability]
	The potential outcome $Y_t(a)$ is conditionally independent of treatment assignment $A_t$, i.e.
	$Y_t(a)\perp A_t |X_t, Z_t, \mathcal{F}_t$.
\end{condition}

\section{Simulation settings}

There are two steps to conducting this simulation study.
First, we must simulate a hypothetical ML algorithm for predicting a patient's risk of readmission.
Second, we simulate data to investigate the operating characteristics of various monitoring procedures.
We discuss each step in turn.

\textbf{Simulating the algorithm.} We consider a setup with only two patient variables $X \in \mathbb{R}{10}$, generated independently from a normal distribution with mean zero and variance 4.
In the pre-deployment setting, we suppose the treatment was assigned uniformly at random.
The data is generated from the logistic regression model
$$
\text{logit}(y=1|x,a) = - 0.5 x_1 - x_2 + 0.5a + x_1 a + 2 x_2 a.
$$
A random forest classifier $\hat{f}$ is trained using 5000 observations and locked thereafter.
The model outputs the predicted risk (probability) of a readmission under a particular treatment.

\textbf{Treatment propensities.}
We suppose that treatment decisions are assigned according to a logistic regression model of the form
$$
\text{logit}(a=1|x,\hat{f}_t) = \beta \left[ \hat{f}_t(x, a=1) - \hat{f}_t(x, a=0))\right]
$$
for some $\beta \in \mathbb{R}$.
We suppose $\beta = -6$ in the observational setting (i.e. clinicians closely follow recommendations from the ML algorithm) and $\beta=-2$ in the interventional setting (i.e. randomization weights favor following the ML algorithm).

\textbf{Shifts in the outcome distribution.}
We simulate shifts in the conditional distribution of $Y(a)$ under treatment $a$ for $a=0$ versus $a=1$ in the following subgroups:
\begin{itemize}
	\item \texttt{Subgroup all}: all $x$
	\item \texttt{Subgroup known}: $x_1 \in (-1,2)$ and $|x_2| < 2.5$ 
	\item \texttt{Subgroup misspec}: $|x_1| < 1.5$ and $|x_2| < 1.5$ 
\end{itemize}
The simulations vary in the magnitude of the increase ($c=10\%$ versus $c=20\%$).
Specifically, for subgroup $\mathcal{S}$ and treatment $a$, the conditional distribution shifts to
\begin{align*}
	p_{1}(Y=1|X,A):=&p_{0}(Y=1|X,A) - \mathbbm{1}\{p_0(Y=1|X,A) > 0.5, X \in \mathcal{S}, A = a\} * c\\
	& + \mathbbm{1}\{p_0(Y=1|X,A) < 0.5, X \in \mathcal{S}, A = a\} * c
\end{align*}
where $p_0$ is the pre-change probability and $p_1$ is the post-change probability.

\textbf{Pre-specified subgroups in the monitoring procedures.}
The subgroups considered by monitoring procedures 2I, 2O, 3I, and 3O are: $\mathcal{S}_1 = \{x : x\in \mathbb{R}\}$,  $\mathcal{S}_2 = \{x : x_1 \in (-1,2), |x_2| < 2.5\}$, and $\mathcal{S}_3 = \mathcal{S}_1\setminus \mathcal{S}_2$.
For computational speed, observations were monitored in batches of 50.

\section{Additional simulation results}

\begin{figure}
	\centering
	\includegraphics[width=0.5\textwidth]{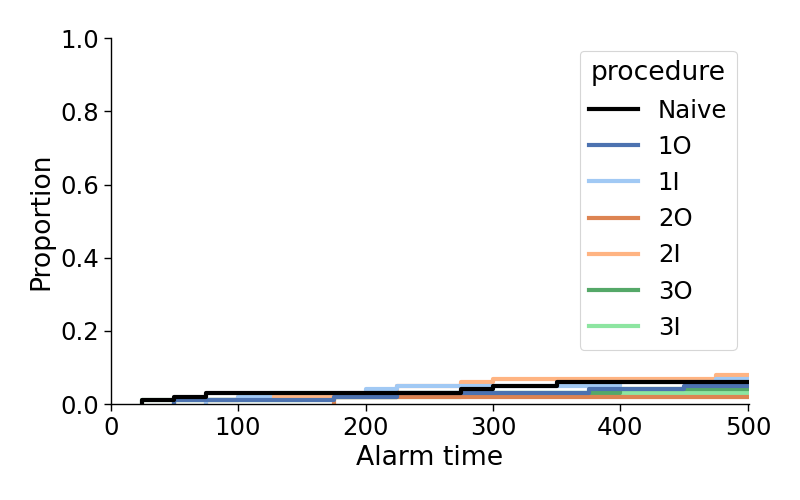}
	\caption{Assessing Type I error for monitoring procedures}
	\label{fig:cusum_type_i}
\end{figure}

We also ran a simulation to verify Type I error control of the monitoring procedures.
The nominal rate was set to $\alpha = 0.1$ and the data was simulated to be IID over time.
Results shown in Figure~\ref{fig:cusum_type_i} are for $\delta = 0.02$.

\begin{figure}
	\centering
	
	\textit{Treatment A=0 shift, Gradual shift up to magnitude 10\%}
	
	\footnotesize{\textit{\qquad \qquad Shifts in All \qquad\qquad \qquad \qquad  Subgroup known  \qquad  \qquad \qquad\qquad  Subgroup misspec}}
	\includegraphics[width=0.3\textwidth]{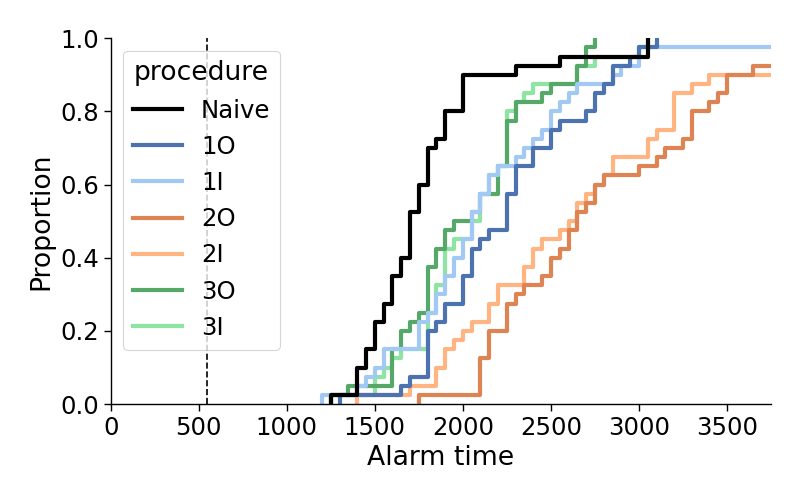}
	\includegraphics[width=0.3\textwidth]{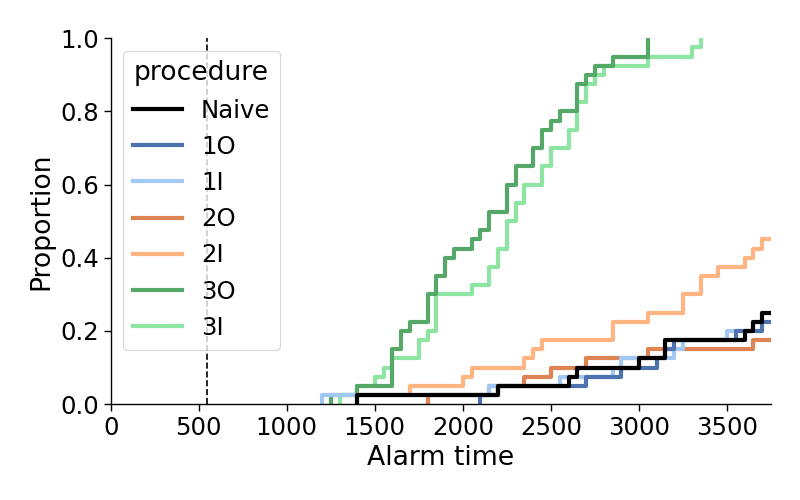}
	\includegraphics[width=0.3\textwidth]{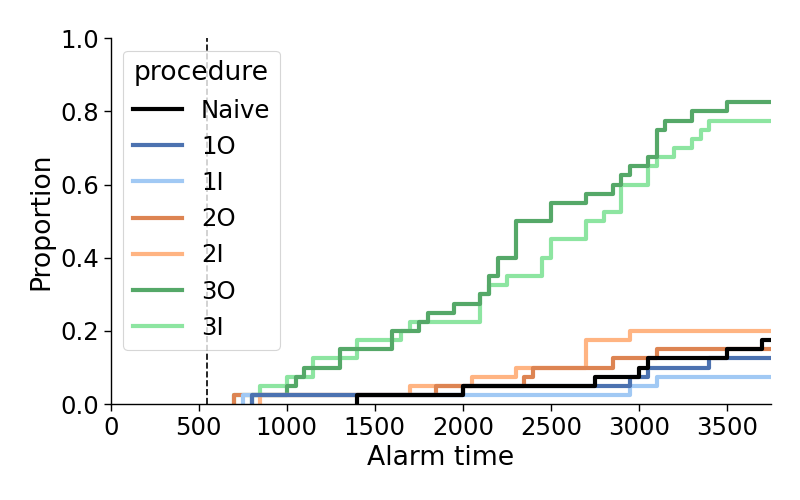}
	
	\textit{Treatment A=1 shift, Gradual shift up to magnitude 10\%}
	
	\footnotesize{\textit{\qquad \qquad Shifts in All \qquad\qquad \qquad \qquad  Subgroup known  \qquad  \qquad \qquad\qquad  Subgroup misspec}}    \includegraphics[width=0.3\textwidth]{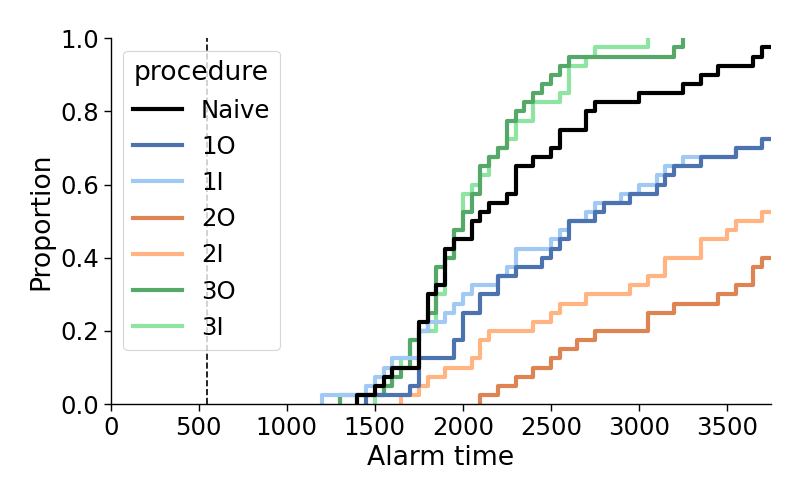}
	\includegraphics[width=0.3\textwidth]{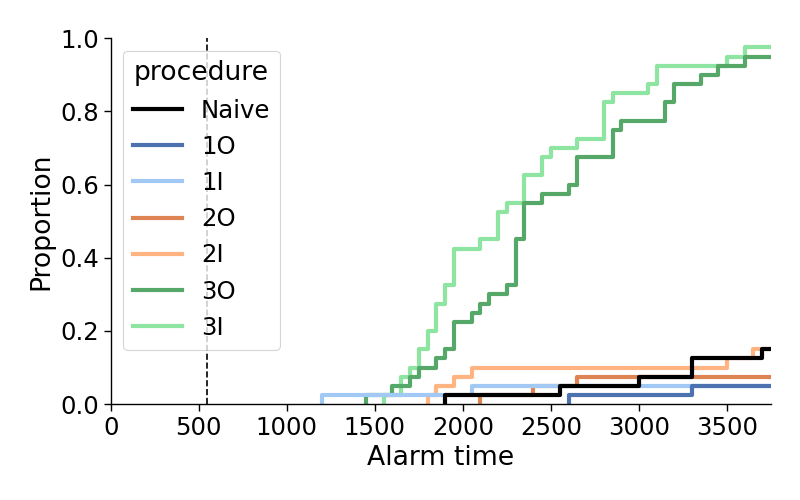}
	\includegraphics[width=0.3\textwidth]{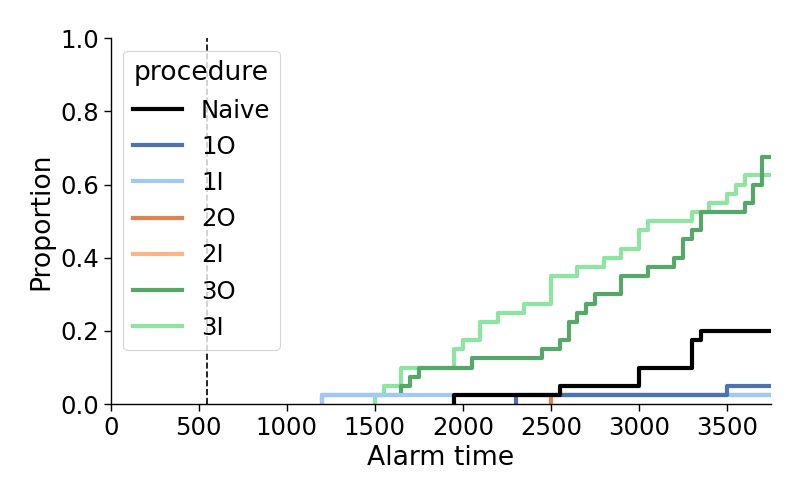}
	
	\textit{Treatment A=0 shift, Gradual shift up to magnitude 20\%}
	
	\footnotesize{\textit{\qquad \qquad Shifts in All \qquad\qquad \qquad \qquad  Subgroup known  \qquad  \qquad \qquad\qquad  Subgroup misspec}}
	\includegraphics[width=0.3\textwidth]{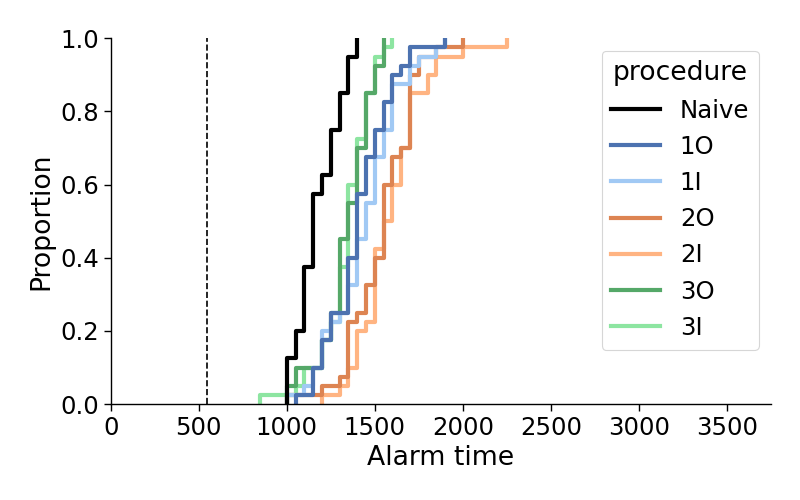}
	\includegraphics[width=0.3\textwidth]{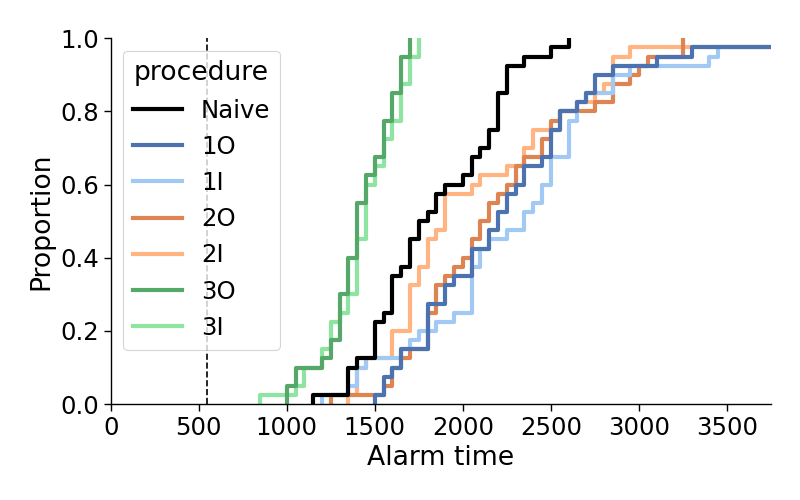}
	\includegraphics[width=0.3\textwidth]{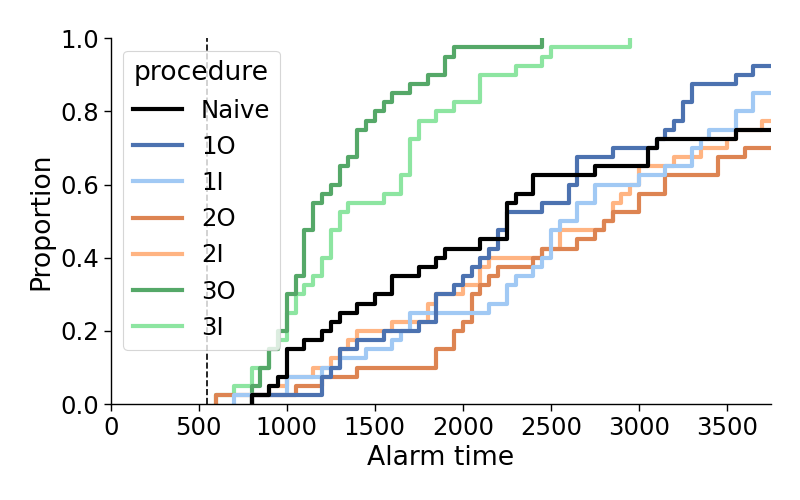}
	
	\textit{Treatment A=1 shift, Gradual shift up to magnitude 20\%}
	
	\footnotesize{\textit{\qquad \qquad Shifts in All \qquad\qquad \qquad \qquad  Subgroup known  \qquad  \qquad \qquad\qquad  Subgroup misspec}}
	\includegraphics[width=0.3\textwidth]{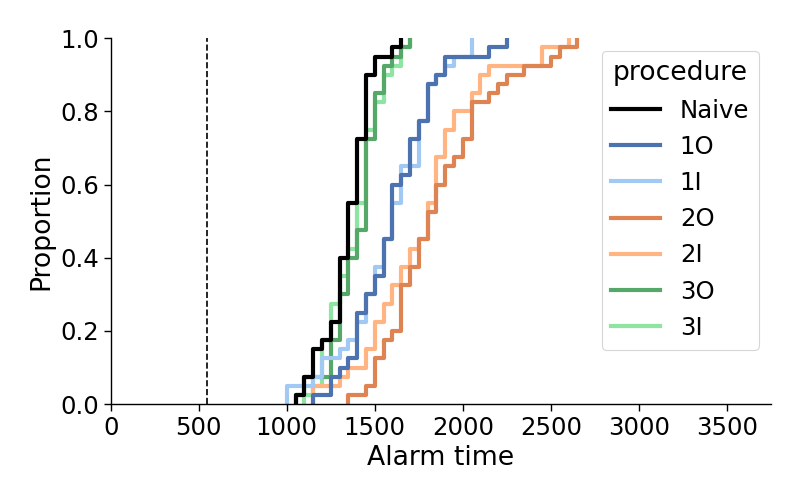}
	\includegraphics[width=0.3\textwidth]{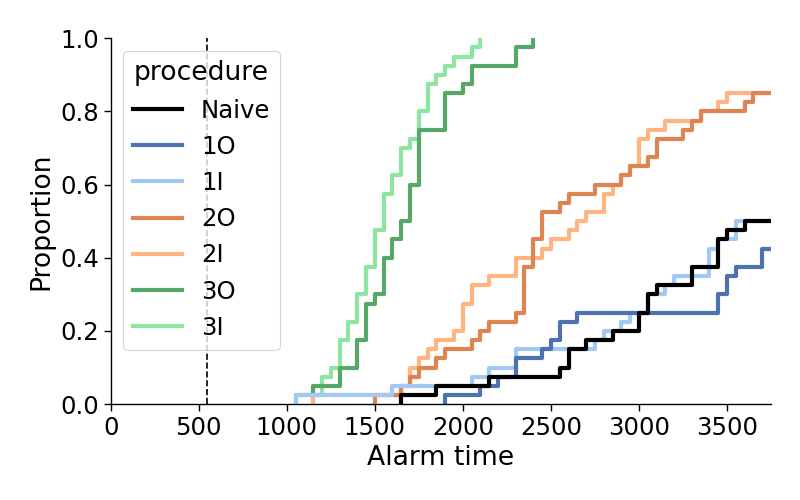}
	\includegraphics[width=0.3\textwidth]{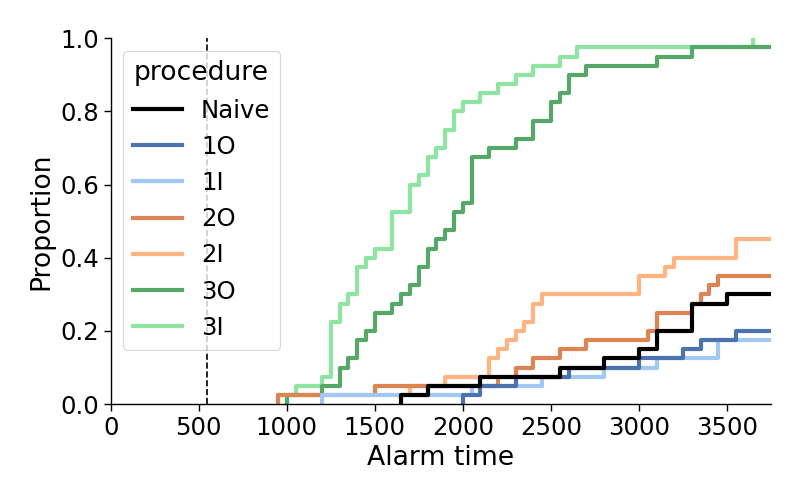}

	\caption{Statistical power of different procedures, as characterized by the proportion of alarms fired at each time point.
		The simulated shift in the conditional distribution of the outcome is gradual over time, with the start time of the shift indicated by the dashed vertical line.}
	\label{fig:cusum_gradual}
\end{figure}

\end{document}